\documentclass[letterpaper, 10 pt, conference]{ieeeconf}  

\IEEEoverridecommandlockouts                              

\usepackage{gensymb}
\usepackage{graphicx} 
\usepackage{times} 
\usepackage{amsmath} 
\usepackage{amssymb}  
\usepackage{url}
\usepackage{cite}
\usepackage{enumerate}
\usepackage{bm}
\usepackage{cases}
\usepackage{mathtools}

\usepackage{multirow}
\usepackage{colortbl}
\usepackage{booktabs} 
\usepackage{makecell}
\usepackage{algpseudocode}

\usepackage[caption=false, font=footnotesize]{subfig}
\usepackage{algorithmicx}
\usepackage{algorithm}
\usepackage{algpseudocode}
\usepackage{booktabs, ragged2e}
\usepackage{tabularx}
\usepackage{xcolor}
\usepackage[para]{threeparttable}
\usepackage{colortbl}
\usepackage{flushend} 
\usepackage{caption}
\usepackage{float}
 
\makeatletter
\let\NAT@parse\undefined
\makeatother
\usepackage[hidelinks]{hyperref} 

\newcolumntype{C}{>{\Centering\arraybackslash}X}
\newcolumntype{L}{>{\raggedright\arraybackslash}X}
\newcolumntype{R}{>{\raggedleft\arraybackslash}X}
\begin{document}
\title{\LARGE \bf
DragTraffic: Interactive and Controllable Traffic Scene Generation for Autonomous Driving


\author{Sheng WANG$^{1}$, Ge SUN$^{1}$, Fulong MA$^{2}$, Tianshuai HU$^{1}$, Qiang QIN$^{3}$, Yongkang SONG$^{4}$, \\ Lei ZHU$^{2}$ and Junwei LIANG$^{2}$}



 
\thanks{This work was supported by Lotus Technology Ltd. through The Hong Kong University of Science and Technology (GZ) under Cooperation Project R00082, Corresponding author: Junwei LIANG.}

\thanks{$^{1}$Sheng WANG, Ge SUN and Tianshuai HU are with Robotics and Autonomous Systems, Division of Emerging Interdisciplinary Areas (EMIA) under Interdisciplinary Programs Office (IPO), The Hong Kong University of Science and Technology, Hong Kong SAR, China. \texttt{\{swangei, gsunah, thuaj\}@connect.ust.hk}} 

\thanks{$^{2}$Junwei LIANG is with Artificial Intelligence Thrust, Fulong MA and Lei ZHU are with Robotics and Autonomous Systems Thrust, The Hong Kong University of Science and Technology (Guangzhou), Guangzhou 511400, China. \texttt{\{junweiliang, leizhu\}@hkust-gz.edu.cn, fmaaf@connect.hkust-gz.edu.cn}}

\thanks{$^{3}$Qiang QIN is with Department of Production Engineering, KTH Royal Institute of Technology, Sweden. \texttt{qiangq@kth.se}}

\thanks{$^{4}$Yongkang Song is with Lotus Technology Ltd, China. \texttt{yongkang.song@lotuscars.com.cn}}




}
\maketitle

\begin{abstract}

Evaluating and training autonomous driving systems require diverse and scalable corner cases. However, most existing scene generation methods lack controllability, accuracy, and versatility, resulting in unsatisfactory generation results. Inspired by DragGAN in image generation, we propose DragTraffic, a generalized, interactive, and controllable traffic scene generation framework based on conditional diffusion. DragTraffic enables non-experts to generate a variety of realistic driving scenarios for different types of traffic agents through an adaptive mixture expert architecture. We employ a regression model to provide a general initial solution and a refinement process based on the conditional diffusion model to ensure diversity. User-customized context is introduced through cross-attention to ensure high controllability. Experiments on a real-world driving dataset show that DragTraffic outperforms existing methods in terms of authenticity, diversity, and freedom. Demo videos and code are available at
\href{https://chantsss.github.io/Dragtraffic/}{\color{blue}{https://chantsss.github.io/Dragtraffic/}.}
\end{abstract}

\renewcommand{\arraystretch}{1.0}
\section{Introduction\label{sect:intro}}

The safety of autonomous driving systems relies heavily on the richness of the dataset scenarios. However, due to various constraints such as safety issues, geographical environment, and weather changes, it is difficult for collected data to cover all situations. This poses challenges for training and evaluating planning and prediction modules, especially for extreme scenarios. To address this, simulators such as SUMO \cite{sumo} and CARLA \cite{carla} have been used to manually set scenarios. While rule-based simulations offer interpretability and viable trajectories without extensive data, they have limitations in accuracy, generalization, and adaptability. Furthermore, they require substantial expert knowledge to establish the necessary rules.

\begin{figure}[t] \centering 
\includegraphics[width=\linewidth]{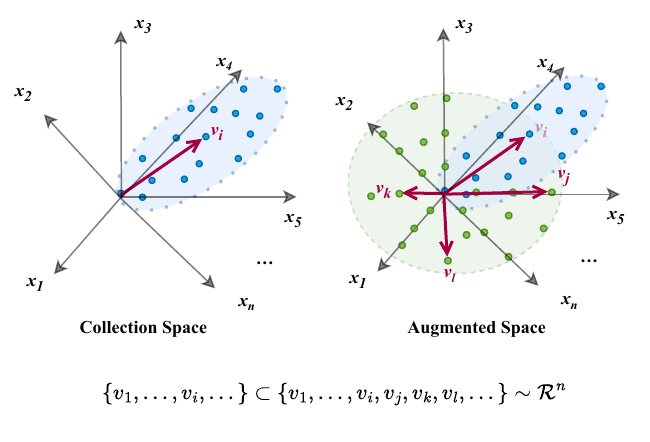} \caption{\textbf{The dataset sample space.} The left image illustrates the distribution of the collected data, while the right image shows the expanded sample space achieved through data augmentation. In this context, $x$ represents the different dimensions that constitute the dataset, and $\bm{v}$ represents the specific samples collected.} \label{fig:first_img} 
\end{figure}

In contrast, data-driven methods have been developed to enable agents in these environments to emulate the behaviors of real traffic participants. The recent Sim Agents Challenge \cite{waymosimagent} based on the Waymo dataset provides a standard benchmark and specifies input and output forms for scene generation tasks. Several works have used learning-based methods to achieve good results, particularly in terms of accuracy \cite{philion2023trajeglish}\cite{qcnet}\cite{mtr}. However, most of these works formulate the scene generation task as a motion prediction task, requiring 10 frames of historical information as input and limiting the freedom of scene construction. As a result, they can only reason about future scenarios based on existing historical trajectories, while ignoring requirements such as scene editing and agent insertion. Other researchers have looked at generating challenging scenes in a more flexible way, such as SceneGen \cite{tan2021scenegen} and TrafficGen \cite{feng2023trafficgen}, which propose building scenes in two stages: vehicle placement and trajectory generation. However, a main shortcoming of these methods is the lack of controllability, which means they cannot ensure expected behavior. This serious problem leads to the generation process being directionless and extremely inefficient when the sample space is large, as shown in the Figure \ref{fig:first_img}. The ideal sample data distribution should be distributed in all dimensions. However, due to factors such as cost and security, high-value data is often scarce. Another problem is that they only focus on vehicles and ignore other types of traffic participants, even though these participants have many interactions. Meanwhile, generative models have been well developed in contentmade great progress in sequence generation tasks such as text, pictures, and videos\cite{text_generation_survey}\cite{ stable_diffusion}\cite{picgeneration2}\cite{videogeneration1}\cite{DiT}. Some prior works are inspired of using generative models to create traffic scenes such as \cite{gan1}, \cite{gan2} utilizing generative adversarial networks (GAN) to generate multiple
trajectories for traffic agents. In order to obtain a more operational trajectory generation method, some researchers try to use conditional diffusion models such as motiondiffuser\cite{motiondiffuser}, and SceneDM \cite{scenedm} \textit{etc}. These studies provide a good foundation for using generative models to create scenes, but there is a general problem. They require significant expert knowledge to establish the complex definition of loss function or post-processing. CTG++\cite{ctg++} provides a idea to introduce the large language model in loss function design in a user friendly way. However, this method requires repeated and complex training processes for different tasks.

In this paper, we propose DragTraffic, an instance behavior level traffic scene generator that is capable of generating realistic and diverse scenes while maintaining a high degree of freedom on controllability. To achieve realism, we employ a regress model to provide initial guesses. To account for the behavioral differences among various traffic agents, we adopt a symmetric hybrid expert architecture that imitates the real behavior of traffic participants on the road from an agent-centric perspective by using a separate model dedicated to the corresponding agent. Inspired by Draggan \cite{draggan}, we adopt the conditional diffusion model to achieve diversity and specific defined context, including position, velocity, heading, length, and width. All these controls can be done through dragging or typing context in an interactive and user-friendly way. Our contributions include:


\begin{itemize}

    \item First, we propose an interactive traffic generation framework, which, to the best of our knowledge, is the first to offer a high degree of freedom for generating and editing traffic scenes.
    
    \item Second, we introduce a solution that utilizes a regression model for initial solutions, a conditional diffusion model for diversity, and a Mixture of Experts (MoE) to accommodate multiple agent types. This approach enables us to generate realistic and diverse traffic scenes with a high level of controllability.
    
    \item Third, we conduct experiments on a real-world driving dataset to evaluate the performance of DragTraffic. The results show that DragTraffic outperforms existing methods in terms of authenticity, diversity, and freedom, among other metrics.   

\end{itemize}

\section{Related Work}
\subsection{Trajectory Prediction}
Current methods for testing and developing autonomous driving systems, such as scenario replay and rule-based approaches, have limitations in accuracy, generalization ability, and adaptive updating. These methods also require a large amount of expert knowledge to build the rules. To address these shortcomings, recent studies have explored deep learning-based motion prediction methods that can model multi-modal traffic scenes\cite{philion2023trajeglish}\cite{mtr}\cite{POP}\cite{mmtransformer}. These methods can be broadly categorized into supervised learning and generative learning approaches.
Supervised learning trains a model with logged trajectories with supervised losses such as L2 loss. One of the challenges is to model inherent multi-modal behavior of the agents. For instance, Trajeglish\cite{philion2023trajeglish} employs the template sets to help the model generate realistic multi-modal trajectories by providing a structured framework for interactions. A series works, MultiPath++\cite{MultiPath++}, DenseTNT\cite{DenseTNT} and GANet\cite{gnet} use static anchors or learned goals to represent the multiple hypothises. GoHome\cite{GOHOME} and YNet\cite{Ynet} predict future occupancy heatmaps, and then decode trajectories from the samples. Many of these approaches use ensembles for further diversified predictions. The next section covers generative approaches. 

\subsection{Generative models for Scene Generation}
Generative models have made significant progress in sequence generation tasks such as text, pictures, and videos. Some prior works have explored the use of generative models to create traffic scenes. For example, GANs have been used to generate multiple trajectories for traffic agents in \cite{gan1} and \cite{gan2}. Variational Autoencoders (VAEs) have been employed in MTG \cite{vae1} and CVAE-H \cite{vae2} to extract representations of historical trajectories of agents and generate future trajectories. CTG \cite{ctg++} uses a conditional diffusion model for generating controllable traffic simulations, while CTG++ \cite{ctg++} introduces a scene-level conditional diffusion model guided by language instructions. MotionDiffuser \cite{motiondiffuser} and SceneDM \cite{scenedm} both use diffusion-based models for predicting multi-agent motion and achieving state-of-the-art results on the Waymo Open Motion Dataset and Waymo Sim Agents Benchmark, respectively. While the above methods have achieved good performance, they often require professionals to design complex optimization constraints and loss functions. In contrast, our proposed framework simplifies the task of scene generation by allowing users to control the generation process through simple drag and click, while ensuring the quality of the generated scenes.

\section{Problem formulation}

Our aim is to generate the expected future motions for agents in a scenario. We adopt a structured vectorized representation to depict the map and agents. The trajectory of a specific agent is denoted as $\bm{\tau}_{0:t} = \{\bm{s}_0,\bm{s}_1,...,\bm{s}_t\}$,  where $\bm{s}_t \in \mathbb{R}^D$ indicating the states including the type, location, heading angle, velocity at time step $t$. The road map is denoted as $\bm{L} = \{\bm{l}_i\}$, where $\bm{l}_i \in \mathbb{R}^{N \times H}$ representing $i_{th}$ lane has $N$ segments and each segment has $H$ lane semantic attributes (\textit{e.g.}, intersections and crosswalks). Specifically, for the task of existing scenario augmentation or editing, we aim to generate $t_F$ steps future trajectories
$$
    \bm{\tau}_{1:t_F} = f({\bm{s}}_{0}^{0}...{\bm{s}}_{0}^{M}, {\bm{s}}_{t_F}^{0},\bm{L}),
$$
where ${\bm{s}_0}^1...{\bm{s}_0}^M$ indicates the initial states of $M$ agents. $\bm{s}_{t_F}^{0}$ represents the condition information. For the task of new scenario creation, we aim to generate $t_F$ steps future trajectories following
$$
    \bm{\tau_}{1:t_F} = f(G(\bm{L}),\bm{s}_{t_F}^{0},\bm{L}),
$$
where $G(\cdot)$ indicates the initial states generation, which can be simply achieved through dragging, typing or an agent placement module.

\section{Methodology\label{subsect:Methodology}}

\begin{figure*}[ht]
    \centering
    \includegraphics[width=\textwidth]{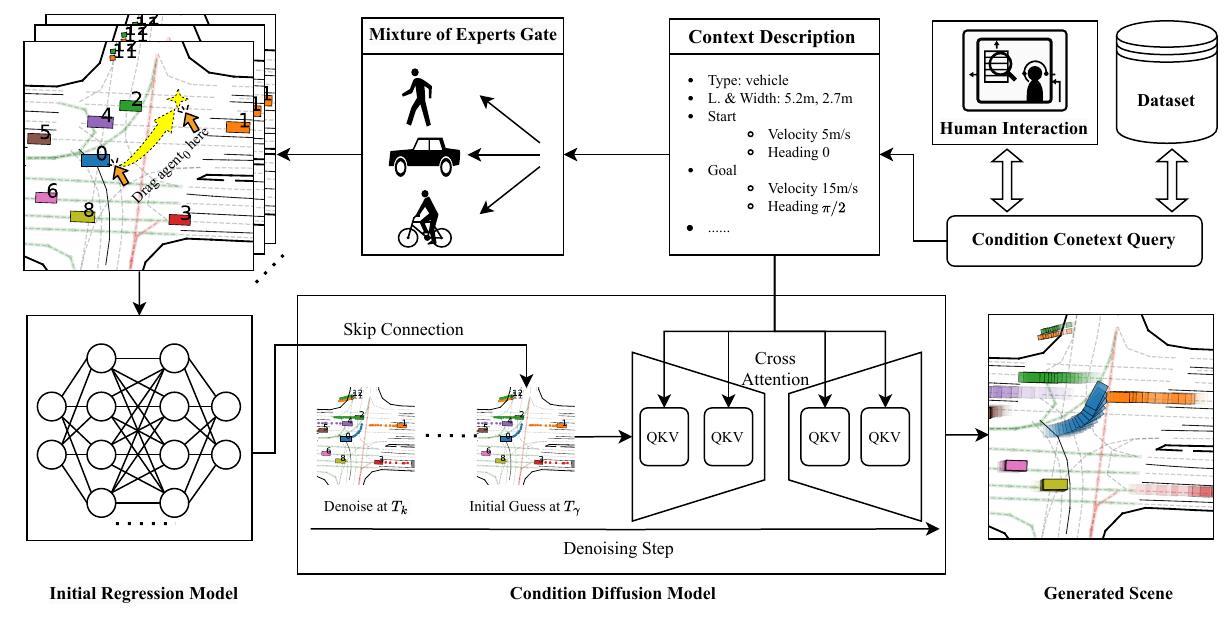}
    \caption{\textbf{The generation pipeline.} 
The Condition Context Query gathers personalized information from the user, either through an interactive UI or by retrieving it from the dataset. The Mixture of Experts Gate selects the appropriate model for inference based on the agent type. Input data is presented as agent-centric vectors. After obtaining the initial solution, it is further refined through diffusion to ultimately generate the scene.}
    \label{fig:overview_structure}
\vspace{-1.5em}
\end{figure*}

\subsection{Conditional Diffusion Model Preliminaries}
Diffusion models consist of a diffusion process that gradually transforms a data distribution into unstructured noise and a reverse process to recover the data distribution\cite{ddpm}. The forward diffusion process acting on $\bm{\tau}_{1:F}$ is defined as
$$
\begin{aligned}
q(\bm{\tau}_{1:F}^{1:k}|{\bm\tau}_{1:F}^{0})& :=\prod_{k=1}^{k}q(\bm{\tau}_{1:F}^{k}|\bm{\tau}_{1:F}^{k-1}),\\
q(\bm{\tau}_{1:F}^{k}|\bm{\tau}_{1:F}^{k-1})& :=\mathcal{N}(\bm{\tau}_{1:F}^{k};\sqrt{1-\beta_{k}}\bm{\tau}_{1:F}^{k-1},\beta_{k}\bm{I}),
\end{aligned}
$$
where the variance schedule $\beta_1,\beta_2,\cdots\beta_k$ is fixed and determines the amount of noise injected at each diffusion step, leading to a gradual corruption of the signal into an isotropic Gaussian distribution. To generate trajectories, we aim to reverse this diffusion process by utilizing a learned conditional denoising model, which is iteratively applied starting from sampled noise. Given the context information $\bm{c}(\bm{\tau}_0, \bm{s}_{t_F}^{0}, \bm{L})$, the reverse diffusion process is
$$
\begin{aligned}
p_{\theta}(\bm{\tau}_{1:F}^{0:k}|\bm{c})& :=p(\bm{\tau}_{1:F}^{k})\prod_{k=1}^{k}p_{\theta}(\bm{\tau}_{1:F}^{k-1}|\bm{\tau}_{1:F}^{k},\bm{c}), \\
p_{\theta}(\bm{\tau}_{1:F}^{k-1}|\bm{\tau}_{1:F}^{k},\bm{c})& :=\mathcal{N}(\bm{\tau}_{1:F}^{k-1};\bm{\mu}_{\theta}(\bm{\tau}^{k},k,\bm{c}),\bm{\Sigma}_{\theta}(\bm{\tau}_{1:F}^{k},k,\bm{c})).
\end{aligned}
$$
The distribution $p(\bm{\tau}_{1:F}^{k})$ is a normal distribution and $\theta$ denotes the parameters of the diffusion model. In this work, we adopt the idea proposed in \cite{led} and use the conditional diffusion model as a refinement module as shown in Figure \ref{fig:overview_structure}.
This is to say, a skip connection is used to generate $\bm{\tau}_{1:F}^0$ following
$$
\begin{aligned}
\bm{\tau}^k_{1:F} \sim \bm{\tau}^{\star}_{1:F} &= f_{\mathrm{init}}(\bm{c}), \\
\bm{\tau}^{\gamma}_{1:F} &= f_{\mathrm{denoise}}(\bm{\tau}_{1:F}^{\gamma+1},\bm{c}),
\gamma=k-1,\cdots,0,
\end{aligned}
$$where $f_{\mathrm{init}}(\cdot)$ is a standard motion forecasting model providing the initial gueses for better regression performance purpose, we will elaborate its effectiveness in following experiments section.

\subsection{Context Description}
Due to the high degree of freedom of our generation scheme, the context description can consist of some or all of the following information: vehicle type, length and width, starting position, starting speed, starting orientation, target position, target speed, target orientation. 


\subsection{Initial Trajectory Generation}
We utilize Multipath++ \cite{MultiPath++} as the initial backbone for trajectory generation, which employs multi-context gating (MCG) blocks. MCG can be seen as an approximation of cross-attention. Instead of having each of the n elements attend to all m elements of the other set, MCG condenses the other set into a single context vector. The prediction heads take $\bm{c}$ then output the future $\bm{\tau}$ and $K$ probabilities. This initial regression module is trained by minimizing the MSE loss of the predicted trajectory which is the closest to the ground truth trajectory.


\subsection{Diffusion Refinement}

Here, we detail the design of the denoising module  $f_{\text {denoise }}(\cdot)$ , which denoises the trajectory  $\bm{\tau}_{1:F}^{\gamma+1}$ conditioned on context $\bm{c}(\bm{\tau}_0, \bm{s}_{t_F}^{0}, \bm{L})$. The denoising module consists of two trainable components: a MCG-based context encoder $f_{\text {context}}(\cdot)$ that learns a social-temporal embedding for the current states and condition states and 
a noise estimation module $f_{\boldsymbol{\epsilon}}(\cdot)$ that estimates the noise to be reduced. The $\gamma_{th}$ denoising step follows
\begin{align}\boldsymbol{\epsilon}_{\theta}^{\gamma} & = f_{\boldsymbol{\epsilon}}\left(\bm{\tau}^{\gamma+1}_{1:F}, f_{\text {context}}(\bm{c}), \gamma+1\right), \\\bm{\tau}^{\gamma}_{1:F} & = \frac{1}{\sqrt{\alpha_{\gamma}}}\left(\bm{\tau}^{\gamma+1}_{1:F}-\frac{1-\alpha_{\gamma}}{\sqrt{1-\bar{\alpha}_{\gamma}}} \boldsymbol{\epsilon}_{\theta}^{\gamma}\right)+\sqrt{1-\alpha_{\gamma}} \mathbf{z},\end{align}
where $\alpha_{\gamma}:=1- \beta_{\gamma}$  and  $\bar{\alpha}_{\gamma}:=\prod_{i=1}^{\gamma} \alpha_{i}$  are parameters in the diffusion process and  $\mathbf{z} \sim \mathcal{N}(\mathbf{z} ; \mathbf{0}, \mathbf{I})$  is a noise.

\subsection{Training Objective}
To achieve the agent type sensive performance, we use a mixture of experts strucure, which includes a gate on the top to switch the suitable model according to the certain agent type. Then the scene generation framework served by symmetric models. Here for simplication purpose, we only illustrate one of them in details. To train a the dragtraffic model, we consider a two stage training strategy, where the first stage trains a denoising module and the second stage focuses on a leapfrog initializer. The reason for using two stages is to make the training more stable. The noise estimation loss is
$$
\mathcal{L}_{\mathrm{NE}}=\|\boldsymbol{\epsilon}-f_{\boldsymbol{\epsilon}}(\bm{\tau}^{\gamma+1}_{1:F},f_{\mathrm{context}}(\mathbf{c}),\gamma+1)\|_2,
$$
 where $\boldsymbol{\epsilon}\sim\mathcal{N}(\boldsymbol{\epsilon};\bm{0},\bm{I})$ and the diffused
 trajectory $\bm{\tau}_{1:F}^{\gamma+1}=\sqrt{\bar{\alpha}_\gamma}\bm{\tau}_{1:F}^{0}+\sqrt{1-\bar{\alpha}_\gamma}\boldsymbol{\epsilon}.$ We backpropagate this loss to update the parameters in the context encoder $f_{\mathrm{context}}(\cdot)$ and the noise estimation module $f_\epsilon(\cdot).$
In the second stage, we optimize the model by employing a trainable initializer model, while the denoising modules remain frozen. The loss function is
\begin{align}
     \mathcal{L} = \quad \mathcal{L} _\mathrm{MSE}+ \mathcal{L} _\mathrm{NLL}.
\end{align}
We employ a distance-based loss to minimize displacement error and use the Negative Log-Likelihood Loss function to optimize scores, both of which are commonly utilized in motion forecasting tasks.


\begin{figure*}[ht]
    \centering
    \includegraphics[width=\textwidth]{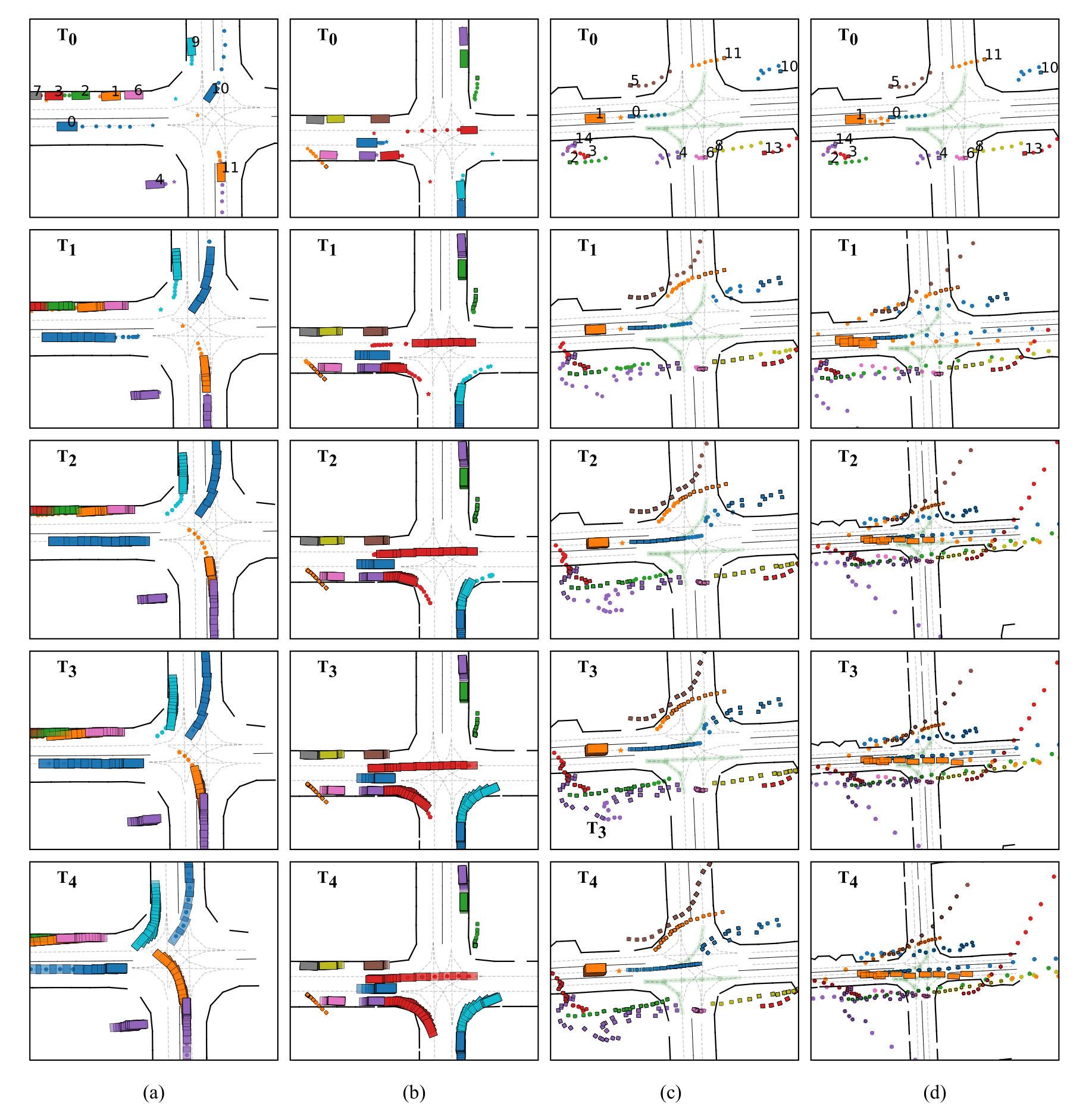}
    \caption{\textbf{The demonstration of creating, editing and correction.} Colored boxes represent agents, with different sizes for each type. A motorcycle is depicted as a long bar, while a pedestrian is represented as a square. A series of colored dots in front of each agent indicates the generated trajectory, and the shadow represents past actions.}
    \label{fig:demo}
\end{figure*}

\section{Experimental Results}\label{sect:experiments}

\subsection{Dataset}
 We utilize the Waymo Open Dataset \cite{waymodataset} to train DragTraffic. It consists of around 70,000 scenarios, each with 20-second traffic information. To optimize the dataset for our purposes, we split each 20-second scenario into 6-second intervals and removed scenarios with less than 32 agents. We then cropped a rectangular area with a 120-meter side length centered on the ego agent and classified scenarios into three datasets: ego-centered, cyclist-centered, and pedestrian-centered. We further filtered out scenarios with less than 30 frames and invalid end points, resulting in 49,884, 29,046, and 9,344 cases, respectively. We then split the remaining cases into training, non-overlapping validation, and test datasets in an 80\%, 10\%, 10\% ratio. To ensure fair comparison with other methods, we benchmarked the trained models on the test set and followed the placement and generation pipeline of TrafficGen, which we considered a robust baseline. Our evaluation produced both quantitative and qualitative results.

\begin{table*}[ht]
  \centering
  \vspace{0.1cm}
  \caption{Scene Generation Quality Evaluation}
  \label{Quality}
\begin{tabular}{|c|c|c|c|c|c|}
\hline
\textbf{Dataset}                      & \textbf{Model}                                     & \textbf{Min ADE$_6$}                   & \textbf{Min FDE$_6$}                  & \textbf{Heading Error}                & \textbf{Speed Error}                  \\ \hline
                                      & TrafficGen                                         & 3.32                                  & 5.41                                  & 0.05                                  & 0.05                                  \\ \cline{2-6} 
                                      & TrafficGen Mixture Training with Condition         & 3.09                                  & 4.40                                  & \textbf{0.04}                         & 0.05                                  \\ \cline{2-6} 
\multirow{-3}{*}{\textbf{Vehicle}}    & \cellcolor[HTML]{EFEFEF}DragTraffic with Condition & \cellcolor[HTML]{EFEFEF}\textbf{2.53} & \cellcolor[HTML]{EFEFEF}\textbf{3.33} & \cellcolor[HTML]{EFEFEF}0.05          & \cellcolor[HTML]{EFEFEF}\textbf{0.03} \\ \hline
                                      & TrafficGen                                         & 6.50                                  & 9.04                                  & 0.10                                  & 0.40                                  \\ \cline{2-6} 
                                      & TrafficGen Mixture Training with Condition         & 2.05                                  & 3.53                                  & 0.02                                  & \textbf{0.36}                         \\ \cline{2-6} 
\multirow{-3}{*}{\textbf{Pedestrian}} & \cellcolor[HTML]{EFEFEF}DragTraffic with Condition & \cellcolor[HTML]{EFEFEF}\textbf{1.59} & \cellcolor[HTML]{EFEFEF}\textbf{2.74} & \cellcolor[HTML]{EFEFEF}\textbf{0.02} & \cellcolor[HTML]{EFEFEF}0.40          \\ \hline
                                      & TrafficGen                                         & 17.05                                 & 23.62                                 & 0.41                                  & 0.31                                  \\ \cline{2-6} 
                                      & TrafficGen Mixture Training with Condition         & \textbf{3.34}                         & \textbf{4.31}                         & \textbf{0.06}                         & \textbf{0.21}                         \\ \cline{2-6} 
\multirow{-3}{*}{\textbf{Cyclist}}    & \cellcolor[HTML]{EFEFEF}DragTraffic with Condition & \cellcolor[HTML]{EFEFEF}3.93          & \cellcolor[HTML]{EFEFEF}6.21          & \cellcolor[HTML]{EFEFEF}0.07          & \cellcolor[HTML]{EFEFEF}0.37          \\ \hline
\end{tabular}
\vspace{-0.1cm}
\end{table*}


\subsection{Metrics}

To evaluate the performance of our framework, we employed two metrics: scenario collision rate (SCR) and motion forecasting related metrics. The SCR measures the consistency of the generated vehicle's behaviors by calculating the average percentage of vehicles that collide with others in each scenario. We consider two vehicles as colliding if their bounding boxes overlap above a predefined IOU threshold. For the open-loop evaluation, we used the common metrics in trajectory prediction tasks: MinADE$_k$, MinFDE$_k$, Heading error and Speed error. These metrics are calculated based on the trajectory with the cloest endpoint to the ground truth over k predictions. Since our condition context includes endpoint information, we treated the point before the last one as the endpoint when calculating the MinFDE.

\subsection{Implementation details}
During the model training phase, we pre-train the conditional diffusion model for 100 epochs and then freeze its parameters. Similar to \cite{led}, after obtaining the initial guess from the regression model, we employ the standard U-net diffusion model to carry out 5 steps of denoising. The total number of denoising steps is 100 for both training and testing. For the Initializer model, we add an MLP layer on top of the backbone to encode the condition information of dimension 8 and obtain hidden features of length 1024, which are then concatenated with the outputs of the state encoder and the lane encoder. We fine-tune the diffusion model and the Initializer model together for 40 epochs. All other baselines that do not utilize the diffusion model are trained for 150 epochs, with the one showing the lowest loss selected for experimental evaluation. All models are trained with a learning rate decay starting at 0.0003 and decaying to 0.0004, using 4 RTX 3090 GPUs.

\begin{table}[hb]
  \centering
  \caption{Interaction Reasoning Evaluation}
  \label{scr}
\begin{tabular}{|c|c|c|}
\hline
\textbf{Rollout Interval} & \textbf{Model}                             & \textbf{SCR(\%)} \\ \hline
\multirow{2}{*}{3s}        & T.G. Mixture Training with Condition & 15.00               \\ \cline{2-3} 
                          & DragTraffic with Condition                 & 9.20              \\ \hline
\multirow{2}{*}{6s}        & T.G. Mixture Training with Condition & 14.43            \\ \cline{2-3} 
                          & DragTraffic with Condition                 & 8.97             \\ \hline
\multirow{2}{*}{9s}        & T.G. Mixture Training with Condition & 13.59            \\ \cline{2-3} 
                          & DragTraffic with Condition                 & 3.33             \\ \hline
\end{tabular}
\end{table}

\subsection{Results and discussion}
We believe that an outstanding scene generator should possess two essential characteristics. Firstly, it must generate realistic and reasonable trajectories by providing accurate context. Similar to the trajectory prediction task, we use displacement error metrics to evaluate its performance. Secondly, a remarkable scene generator should offer a high degree of freedom and controllability, enabling users to create realistic scenes while exploring rare and high-value corner cases. This is where DragTraffic excels compared to other existing frameworks. In the following sections, we will elaborate on the experimental results, focusing particularly on these two characteristics.

\subsubsection{Scene Generation}

To assess the diversity of scenario generation, we conducted three subsets of tests on MinADE and MinFDE for different agent types. The results, presented in the Table \ref{Quality}, demonstrate that our approach achieves good predictive performance for various agents and can approximate the ground truth. In contrast, plain TrafficGen performs well for vehicle agents but exhibits significant performance degradation for pedestrians and bicycles due to its reliance on vehicle data, which oversimplifies the problem and reduces model generalizability. To ensure fairness, we trained trafficGen on a mixed dataset, which also yielded promising results on some metrics. This highlights the importance of establishing a framework for different agent types. DragTraffic outperforms other models on both vehicle and pedestrian datasets, thanks to its MoE structure, which enhances its generalizability. However, we observe that DragTraffic does not perform well on the Cyclist dataset, possibly because the proportion of cyclists in the scene data is too small.
In Table \ref{scr}, we present the results of the Interaction Reasoning Evaluation with sampling intervals of 3s, 6s, and 9s. We collected 300 scenes with high interactivity between different agents for evaluation, including 200 scenes related to pedestrians and 100 scenes related to cyclists. The performance of DragTraffic under the three sampling conditions is superior to the baseline. Interestingly, we observe that the SCR varies with the sampling interval, which is contrary to the results obtained in TrafficGen\cite{feng2023trafficgen}. However, we believe this is reasonable because multiple rollouts can introduce more cumulative errors.

    \subsubsection{Scene Editing \& Inpaiting}
Fig. \ref{fig:demo} demonstrates the quality of scenes generated by DragTraffic based on existing data. We use the information of current frame and the 90th frame in the dataset as conditions. The left-turn and right-turn scenes at the intersection where the most interactions occur are respectively shown in columns (a) and (b). These scenes reflect courtesy and competition for right of way among different agents. For example, in (a), Agent 9 and Agent 11 engage in a fierce competition for the right of way, while Agent 0 gives way. This shows that DragTraffic can simulate traffic participants in different situations under highly dynamic and complex traffic intersections to make reasonable, smooth, and realistic future actions, reflecting real-world characteristics. Next, we verify that this capability is not limited to simple log replay in more complex scenarios (c) and (d). We focus on agent 1, which is waiting at an intersection for the motorcycle in front to start. However, the movement of a large number of pedestrians around interferes with the decision-making of agent 1, which has a conservative driving style. As a result, it continues to watch the movements of pedestrians even when the motorcycle is already driving to the intersection, leading to the phenomenon of robot freezing, which is common in the fields of autonomous driving and robotics. To address this issue, we design similar scenarios by setting the condition information for agent 1 (100 meters in front of it, longitudinal speed of 20m/s, and latitudinal speed of -2m/s) to encourage more proactive behavior. As shown in (d), the results generated by DragTraffic demonstrate effective control over scene editing. Note that we only show the control of agent 1 for the convenience of explanation, but in fact, DragTraffic allows us to control the generation of multiple agents simultaneously. Unlike other generation control methods that rely on complex optimization constraints and loss designs, our generation processes require only simple dragging or typing interactions, showcasing the superiority of this framework.

\section{Future Work\label{sect:future_work}}

While our proposed framework can generate realistic and diverse traffic scenarios, there are still areas for improvement. For example, we can implement post-processing or sampling techniques to ensure that the generated driving actions adhere to dynamic constraints. Another promising avenue for future work is to develop a multi-round generation process incorporating human feedback. 
\section{CONCLUSIONS \label{sect:conclusion}}

In this paper, we introduced DragTraffic, a generalized, interactive, and controllable traffic scene generation framework based on conditional diffusion. Our framework addresses the limitations of existing scene generation methods in terms of controllability, accuracy, and versatility, enabling non-experts to create a variety of realistic driving scenarios for different types of traffic agents. We achieve this using an adaptive mixture expert architecture with a regression model for initial solutions, followed by refinement through the conditional diffusion model to ensure diversity. The denoising process incorporates user-customized context via cross-attention, enhancing controllability. The qualitative and quantitative results on a real-world driving dataset demonstrate the effectiveness of DragTraffic, which can significantly aid in the evaluation and training of autonomous driving systems by providing diverse and scalable corner cases.




\bibliographystyle{IEEEtran}
\bibliography{IEEEabrv,ref}

\end{document}